\documentclass[conference]{IEEEtran}
\IEEEoverridecommandlockouts
\usepackage{cite}
\usepackage{amsmath,amssymb,amsfonts}
\usepackage{algorithmic}
\usepackage{graphicx}
\usepackage{subcaption}
\usepackage{textcomp}
\usepackage{xcolor}
\usepackage{array}
\usepackage{multirow}
\usepackage{hyperref}
\usepackage{algorithm}

\usepackage{eso-pic}

\def\BibTeX{{\rm B\kern-.05em{\sc i\kern-.025em b}\kern-.08em
    T\kern-.1667em\lower.7ex\hbox{E}\kern-.125emX}}

\begin{document}
\AddToShipoutPictureBG*{%
  \AtPageUpperLeft{%
    \setlength\unitlength{1in}%
    \hspace*{\dimexpr0.5\paperwidth\relax}
    \makebox(0,-0.75)[c]{\textit{Accepted for publication at the 9\textsuperscript{th} IEEE International Conference on Robotics and Automation Engineering (IEEE ICRAE 2024)}}
}}

\title{An Open-source Sim2Real Approach for Sensor-independent Robot Navigation in a Grid

\thanks{\IEEEauthorrefmark{1} Equal Contributions.}

\thanks{This work was developed as part of the ME-599: Machine Learning Control course project and supported by the Department of Mechanical Engineering, University of Washington, Seattle, WA, USA.}
}

\author{
    \IEEEauthorblockN{
        Murad Mehrab Abrar 
        \IEEEauthorrefmark{1},
        Souryadeep Mondal 
        \IEEEauthorrefmark{1},
        and Michelle Hickner 
    }
    \IEEEauthorblockA{
        \textit{Department of Mechanical Engineering, University of Washington, Seattle, WA, USA}\\
        \{mabrar, somondal, mhickner\}@uw.edu
    }
}
\maketitle


\begin{abstract}
This paper presents a Sim2Real (Simulation to Reality) approach to bridge the gap between a trained agent in a simulated environment and its real-world implementation in navigating a robot in a similar setting. Specifically, we focus on navigating a quadruped robot in a real-world grid-like environment inspired by the Gymnasium Frozen Lake --- a highly user-friendly and free Application Programming Interface (API)  
to develop and test Reinforcement Learning (RL) algorithms. We detail the development of a pipeline to transfer motion policies learned in the Frozen Lake simulation to a physical quadruped robot, thus enabling autonomous navigation and obstacle avoidance in a grid without relying on expensive localization and mapping sensors. The work involves training an RL agent in the Frozen Lake environment and utilizing the resulting Q-table to control a 12 Degrees-of-Freedom (DOF) quadruped robot. In addition to detailing the RL implementation, inverse kinematics-based quadruped gaits, and the transfer policy pipeline, we open-source the project on GitHub and include a demonstration video of our Sim2Real transfer approach. This work provides an accessible, straightforward, and low-cost framework for researchers, students, and hobbyists to explore and implement RL-based robot navigation in real-world grid environments. 








\end{abstract}

\begin{IEEEkeywords}
Sim2Real, Quadruped, Robotics, Sensor-independent Navigation, Grid Navigation, Gymnasium, Q-Learning, Reinforcement Learning, Arduino, Policy Transfer.
\end{IEEEkeywords}

\section{Introduction}

Robotics research has increasingly relied on simulated environments to develop, test, and optimize autonomous systems before deploying them in the real world. A challenge in this field is the Sim2Real (Simulation to Reality) gap, where models trained in simulation often fail to perform as well in real-world environments due to differences between simulated and physical realities \cite{hofer2021sim2real, yu2019sim}. Assessing this gap is essential for deploying autonomous systems in applications such as navigation, industrial automation, and robotic manipulation. 

The Sim2Real challenge in robotics has driven significant research efforts aimed at introducing new techniques to narrow this gap. Recent advancements include approaches such as bridging the Sim2Real visual gap using natural language \cite{yu2024natural}, benchmarking the Sim2Real gap in cloth manipulation \cite{blanco2024benchmarking}, addressing the Sim2Real gap in robotic 3-D object classification \cite{weibel2019addressing}, and more discussed in the related work section. 
Despite these advancements, many solutions necessitate complex setups or expensive equipment, limiting their accessibility to freshman students, hobbyists, and new researchers. To effectively bridge the Sim2Real gap in robotics, more research is required to develop simple and accessible solutions that can be easily implemented by a broader range of users.

In this work, we present such a solution by using the Gymnasium Frozen Lake environment \cite{gymnasiumfrozenlake} --- an accessible and free platform with a low learning curve to develop and test Reinforcement Learning (RL) algorithms. The Frozen Lake environment features a simple simulated grid layout where an agent is trained to avoid the holes and navigate autonomously, making it ideal for robotics applications. We adopted this environment to address a real-world sensor-independent navigation challenge in robotics. Our method involves training an RL agent in Frozen Lake and transferring the learned policy to a physical 12 Degrees-of-Freedom (DOF) quadruped robot to navigate within a grid autonomously without using sensors. 
Our work aims to:

\begin{itemize}
    \item Develop and implement a straightforward pipeline that transfers policies learned in the Frozen Lake environment to a physical robot.
    
    \item Create a simple, low-cost, and open-source quadruped robot that avoids obstacles and navigates in a grid using the learned policies without relying on sensors or additional real-world training.

\end{itemize}

The motivation behind choosing the Frozen Lake environment is: (i) It is a well-known RL benchmark with a low learning curve, and (ii) It offers a grid-like environment that can represent many real-world application scenarios, such as navigation within a warehouse, manufacturing facilities, and airports. 
The pipeline developed in this paper involves training an RL agent in simulation, creating a Q-table, transforming the Q-table into coordinates, and writing 12 DOF gait scripts to convert the coordinates into robot movement. 
Additionally, all hardware designs and software are open-sourced on GitHub \footnote{Opensource Link: \href{https://github.com/mehrab-abrar/Sim2Real}{https://github.com/mehrab-abrar/Sim2Real}} to facilitate full replication and further research and a demonstration video is available on YouTube \footnote{Video of the Robot: \href{https://www.youtube.com/watch?v=dDKQaN_zsvU}{https://www.youtube.com/watch?v=dDKQaN\_zsvU}}. This work represents, to the best of our knowledge, the first implementation of Sim2Real robot navigation using the Frozen Lake environment. 




\begin{figure*}[htbp]
\centerline{\includegraphics[width=18cm]{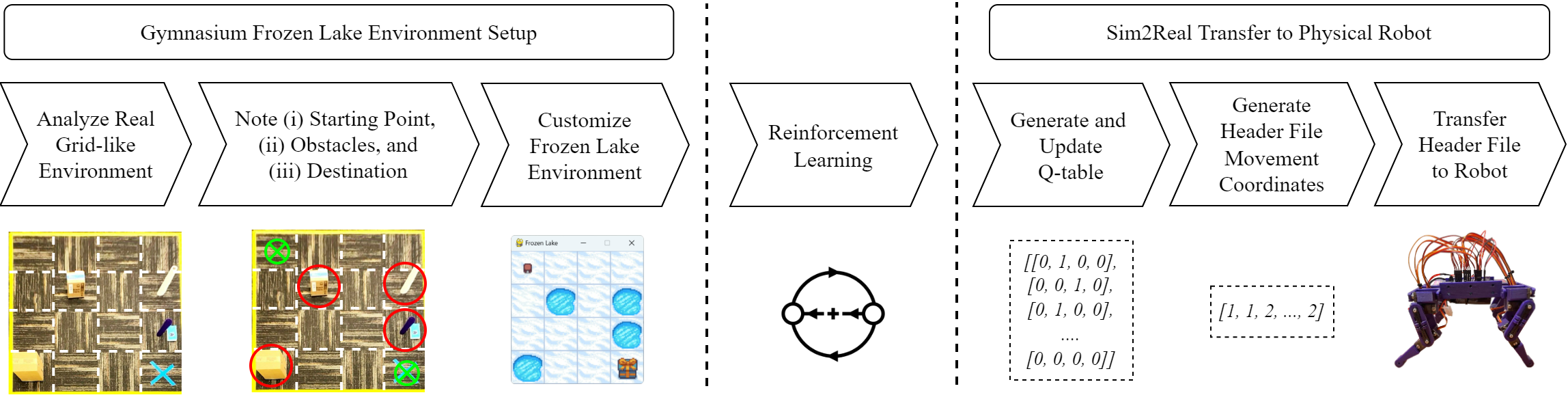}}
\caption{Gymnasium Frozen Lake-inspired Sim2Real architecture for grid navigation}
\label{architecture}
\end{figure*}

\section{Background and Related Work}\label{background}

\subsection{Sim2Real}

Sim2Real, short for “Simulation to Reality”, is a concept in robotics, artificial intelligence (AI), and machine learning that focuses on transferring skills, knowledge, or models learned in a simulated environment to real-world applications \cite{oren2010simulation, daza2023sim}.

\subsection{Q-learning}

Q-learning \cite{watkins1992q} is a model-free RL algorithm that enables agents to learn optimal actions in Markovian environments through interaction, without the need to construct a model of the domain. The expected utility of taking a particular action in a given state is stored in a Q-table, which is updated according to the following equation \cite{rummery1994line}:

\begin{equation}\label{Q-learning equation}
    \begin{aligned}
        Q_{new}(s, a) &= Q_{current}(s, a) + \\
        & \alpha \Big[ R(s, a) + \lambda Q_{max}(s, a) - Q_{current}(s, a) \Big]
    \end{aligned}
\end{equation}

where $Q_{new}(s, a)$ is the new Q-value for the current state $s$ and action $a$, $Q_{current}(s, a)$ is the current Q-value, $\alpha$ is the learning rate, $R(s, a)$ is the reward for taking an action in a state, $\lambda$ is the discount rate, and $Q_{max}(s, a)$ is the maximum expected future reward.

\subsection{Gymnasium Frozen Lake Environment}

Gymnasium is a Python-based open-source toolkit for RL experiments that features a variety of environments. The Frozen Lake environment serves as the foundational simulation for this study. It is a classic RL environment where an agent navigates a grid-based map representing a frozen lake. The goal is to move from a starting point to a destination while avoiding holes that can cause the agent to fall and terminate the episode. The environment is stochastic, meaning that actions may not always result in the intended movement due to the "slippery" nature of the frozen surface. The agent's task is to learn an optimal policy that maximizes the likelihood of reaching the goal while minimizing the risk of falling into holes using a model-free algorithm like Q-learning.
The environment is modeled as an $n \times n$ matrix, with each cell representing a unique state characterized by: State \textbf{S:} Starting point; State \textbf{F:} Frozen surface (safe to walk on); State \textbf{H:} Hole (falls and terminates episode); and State \textbf{G:} Goal.
The state space of Frozen Lake consists of the grid cells that the agent can occupy, and the Action Space consists of the possible actions of the agent, typically \textit{Left, Down, Right, Up}. 
The agent receives a reward of +1 for reaching the goal state and 0 otherwise. Falling into a hole results in the end of the episode with no reward. 
The agent is trained over multiple episodes where it interacts with the environment and updates the Q-values based on the reward structure.

\subsection{Related Work}

Many studies in the literature introduced various techniques for the effective transfer of simulated results to real hardware \cite{zhao2020sim, salvato2021crossing, ju2022transferring}. Yu et al. \cite{yu2019sim} presented a novel approach to transfer bipedal dynamic locomotion control policies from simulation to Darwin OP2 humanoid robot hardware by performing system identification of model parameters in two stages, pre-policy learning and post-policy learning. Hwangbo et al. \cite{hwangbo2019learning} introduced a method for training a neural network policy in simulation and applied it to the ANYmal quadruped robot platform. Andrychowicz et al. \cite{andrychowicz2020learning} used RL to learn dexterous in-hand manipulation policies trained in a simulated environment that can perform vision-based object reorientation on a physical Shadow Dexterous Hand robot. Backhouse et al. \cite{backhouse2022gym2real} introduced introduced Gym2Real, a two-wheeled robotic platform that implements RL by training balance policies in Isaac Gym, a simulation environment accessible to hobbyists.

Bridging the Sim2Real gap in robotic navigation tasks has also been studied by several researchers. Hu et al. \cite{hu2021sim} developed a novel Sim2Real pipeline for a mobile robot to learn to navigate in real-world 3D rough terrains, where the 3D map of the real environment is created in Robot Operating System (ROS) and transferred to a mobile robot equipped with 3D LiDAR, and Stereo camera, and Inertial Measurement Unit (IMU) sensors. Kang et al. \cite{kang2019generalization} integrated simulated and real-world data into deep reinforcement learning for vision-based autonomous navigation and obstacle avoidance for flying robots using a monocular camera. Zhu et al. \cite{zhu2021rule} proposed a rule-based RL algorithm for efficient robot navigation with space reduction, where the navigation maps are built using a Simultaneous Localization and Mapping (SLAM) mobile robot with sensors.


The majority of current Sim2Real approaches in robotics either rely on sensors such as camera feedback, IMU measurements, and LiDARs for navigation, localization, and mapping, or require the use of environments like ROS and Gazebo, which have a steep learning curve. In contrast, we propose a Sim2Real robot navigation approach that is straightforward to learn, easy to implement without compromising functionality, does not require real hardware training, and is accessible to users with varying levels of experience.




\section{Methodology}\label{methodology}

The Sim2Real approach in this work has four main components: (i) Gymnasium Frozen Lake Environment Setup, (ii) Reinforcement Learning, (iii) Sim2Real Transfer to Physical Robot, and (iv) Quadruped Robot Modeling. The architecture of the approach is illustrated in Fig. \ref{architecture}. 


\subsection{Gymnasium Frozen Lake Environment Setup}\label{AA}

The process begins by analyzing the real grid-like environment where the robot is expected to navigate. Critical elements in this environment, such as the Starting Point (S), Obstacles or Holes (H), and Destination or Goal (G) are identified and the Frozen Lake environment is customized to reflect these elements. This customized Frozen Lake environment serves as the training ground for the RL agent, where it learns to find the shortest path and navigate the grid.

\subsection{Reinforcement Learning}

In our implementation, we train an agent using Q-learning due to its simplicity and effectiveness in discrete action spaces, making it suitable for grid-based navigation tasks. The goal of the agent is to learn an optimal policy that allows it to traverse the grid from the Starting Point (S) to the Destination (G) while avoiding Obstacles or Holes (H). 


\subsection{Sim2Real Transfer to Physical Robot}

After the RL agent is trained in the custom Frozen Lake environment, a Q-table is generated and subsequently updated with the optimal policy. From the updated Q-table, movement coordinates are generated and converted into a header file that contains a sequence of commands defining the movement of the robot. This header file is then transferred to the robot code to generate movements and navigate in the real environment. 

This process ensures that the policy learned in the simulation can be effectively applied to the physical robot. The robot is expected to follow the same sequence of movements it learned during the simulation, which negates the necessity of using sensors to detect and avoid obstacles.





\subsection{Modeling the Quadruped}

\begin{figure}[ht]
\centerline{\includegraphics[width=8.5cm]{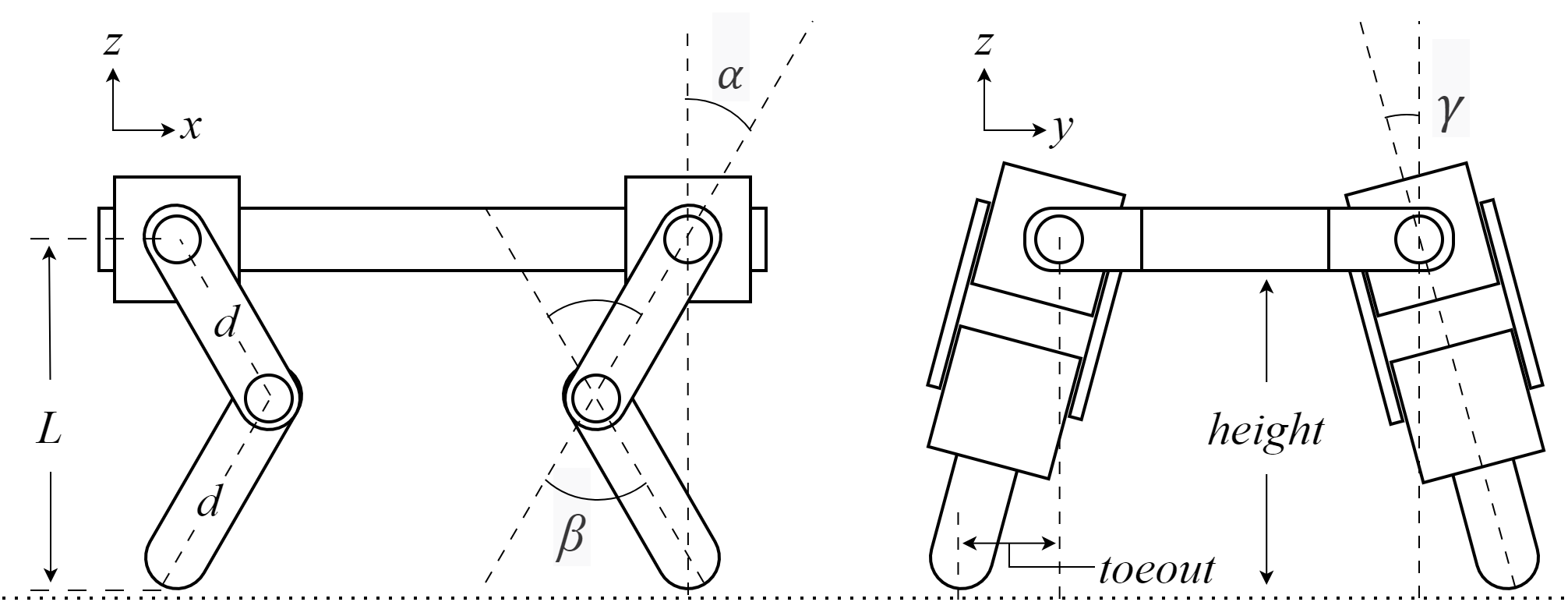}}
\caption{Quadruped robot kinematics}
\label{kinematics}
\end{figure}

\subsubsection{Kinematics and Joint Angle Calculations}

A set of equations calculates the necessary joint angles based on the desired foot trajectories of the robot. For each leg, the kinematic structure is defined by three primary joint angles: hip $\alpha$, knee $\beta$, and ankle $\gamma$, as shown in Fig. \ref{kinematics}. 
The $\alpha$ angle is responsible for the forward and backward leg swing. The $\alpha$ angles for the front and rear legs are calculated using the following equations:

\noindent
\begin{equation}
    \alpha_{\text{front}} = \frac{\beta}{2} - \arcsin\left(\frac{\Delta x}{L}\right)
\end{equation}
\noindent
\begin{equation}
    \alpha_{\text{rear}} = \frac{\beta}{2} + \arcsin\left(\frac{\Delta x}{L}\right)
\end{equation}

Where $\Delta x$ denotes the longitudinal movement along the x-axis and $L$ denotes the leg length. The knee joint angle $\beta$ is derived from the calculated leg length using the following equation:

\noindent
\begin{equation}
    \beta = 2 \times \arccos\left(\frac{L}{2 \times d}\right)
\end{equation}

Where $d$ denotes the length of the upper or lower leg joints, as both upper and lower legs have the same length in our design. The $\gamma$ angle controls the outward or inward rotation of the foot. For the left and right legs, the $\gamma$ angle is computed using the following equations:

\begin{equation}
    \gamma_{\text{left}} = \arctan\left(\frac{\text{toeout}_0 + \Delta y}{\text{height}_0 - \Delta z}\right) - \gamma_0
\end{equation}

\begin{equation}
    \gamma_{\text{right}} = \gamma_0 - \arctan\left(\frac{\text{toeout}_0 - \Delta y}{\text{height}_0 - \Delta z}\right)
\end{equation}

where, $\text{toeout}_0$ is the initial toe-out distance, $\text{height}_0$ is the initial standing height of the robot, $\gamma_0$ is the neutral $\gamma$ position, and $\Delta y$ and $\Delta z$ denote the changes in lateral and vertical movements along the y and z axes, respectively. 
The leg length is determined based on the position of the foot relative to the body. The left and right leg lengths are calculated using the following equations:
\noindent
\begin{equation}
    L_{\text{left}} = \sqrt{\left(L_0 - \frac{\Delta z}{\cos(\gamma_0 + \gamma)}\right)^2 + \Delta x^2}
\end{equation}

\begin{equation}
    L_{\text{right}} = \sqrt{\left(L_0 - \frac{\Delta z}{\cos(\gamma_0 - \gamma)}\right)^2 + \Delta x^2}
\end{equation}

where $L_0$ is the initial leg length.

\subsubsection{Gait Algorithm}


Four walking gaits: \textit{forward creep, reverse creep, left turn,} and \textit{right turn,} along with two additional sharp turn gaits: \textit{left turn 90 degrees} and \textit{right turn 90 degrees} were developed in C++. Each forward function advances the robot by approximately 5 cm, while each turn function changes the yaw angle by 10 degrees. The creeping gait of the robot alternates the foot movements to maintain stability. The algorithm begins by positioning the robot's body to lean toward the direction of movement, followed by sequential foot movements. For instance, during a forward gait:

\begin{itemize}
    \item The body shifts slightly right, positioning itself to move the left legs.
    \item The left legs step forward, followed by the right legs, while the body shifts left.
    \item The robot alternates between shifting the body and moving the feet in sync to maintain balance.
\end{itemize}

The movement is governed by a time-stepped servo control algorithm as follows: 
\noindent
\begin{equation}
    \text{servo speed} = \frac{\theta_{\text{new}} - \theta_{\text{old}}}{\text{timestep}}
\end{equation}

where $\theta_{\text{new}}$ and $\theta_{\text{old}}$ are the new and current joint angles, respectively. The gait sequence can be modified for turning by adjusting the foot positions along an arc, based on a turn angle $\phi$ according to the following equation: 
\noindent
\begin{equation}
    \text{turn position}(f) = \text{body radius} \times \cos\left(\phi_0 + \frac{f \times \pi}{180}\right)
\end{equation}

Using the above mathematical model of the quadruped, the Sim2Real approach is implemented in the experiment section. Algorithm \ref{algo} outlines a pseudocode of the approach. 


\begin{algorithm}
\caption{Sim2Real Robot Navigation with Q-learning}\label{algo}

\textbf{Initialization:}

\begin{algorithmic}[1]
    \STATE Initialize Frozen Lake environment \texttt{env} (4 $\times$ 4 grid)
    \STATE Create a 16$\times$4 Q-table filled with 0's (16 states, 4 actions)\\
    \STATE Set parameters: $\epsilon_{min}$, $\epsilon_{max}$, $LR$, $DR$, $Episodes$, $Steps$
\end{algorithmic}

\textbf{Training the Agent in Frozen Lake:}

\begin{algorithmic}[1]

    \FOR{each episode} 
    \STATE Reset \texttt{env} to initial state + Initialize cumulative reward\\
        \FOR{each step in the episode}
            \STATE Generate random threshold (0 to 1) to decide action:\\
            \IF{random threshold $>$ exploration rate}
                \STATE Choose max Q-value action \textbf{else} Choose random action
            \ENDIF
            \STATE Update Q-table
            \STATE Accumulate reward for episode to total reward
        \ENDFOR
        \STATE Decay exploration rate
    \ENDFOR \\
    
    \STATE Generate movement sequence from Q-table 
    \STATE Export sequence as header file \texttt{data\_array}
\end{algorithmic}

\textbf{Policy Transfer to Physical Robot:}

\begin{algorithmic}[1]

    
    \STATE Import movement sequence from \texttt{data\_array}
    \STATE Generate movement according to sequence using movement functions
    

\end{algorithmic}
\end{algorithm}

\section{Experiment}\label{implementation}

\subsection{Quadruped Hardware Development}

The hardware design of the quadruped robot is inspired by our prior work \cite{abrar2020design}. The design features a point-foot quadruped configuration which is chosen for its effectiveness in achieving stable and precise movements. A total of 27 parts have been designed using OnShape Computer-Aided Design (CAD) software and printed using PrusaSlicer 3D printer.

\begin{figure}[htbp]
\centerline{\includegraphics[width=8cm]{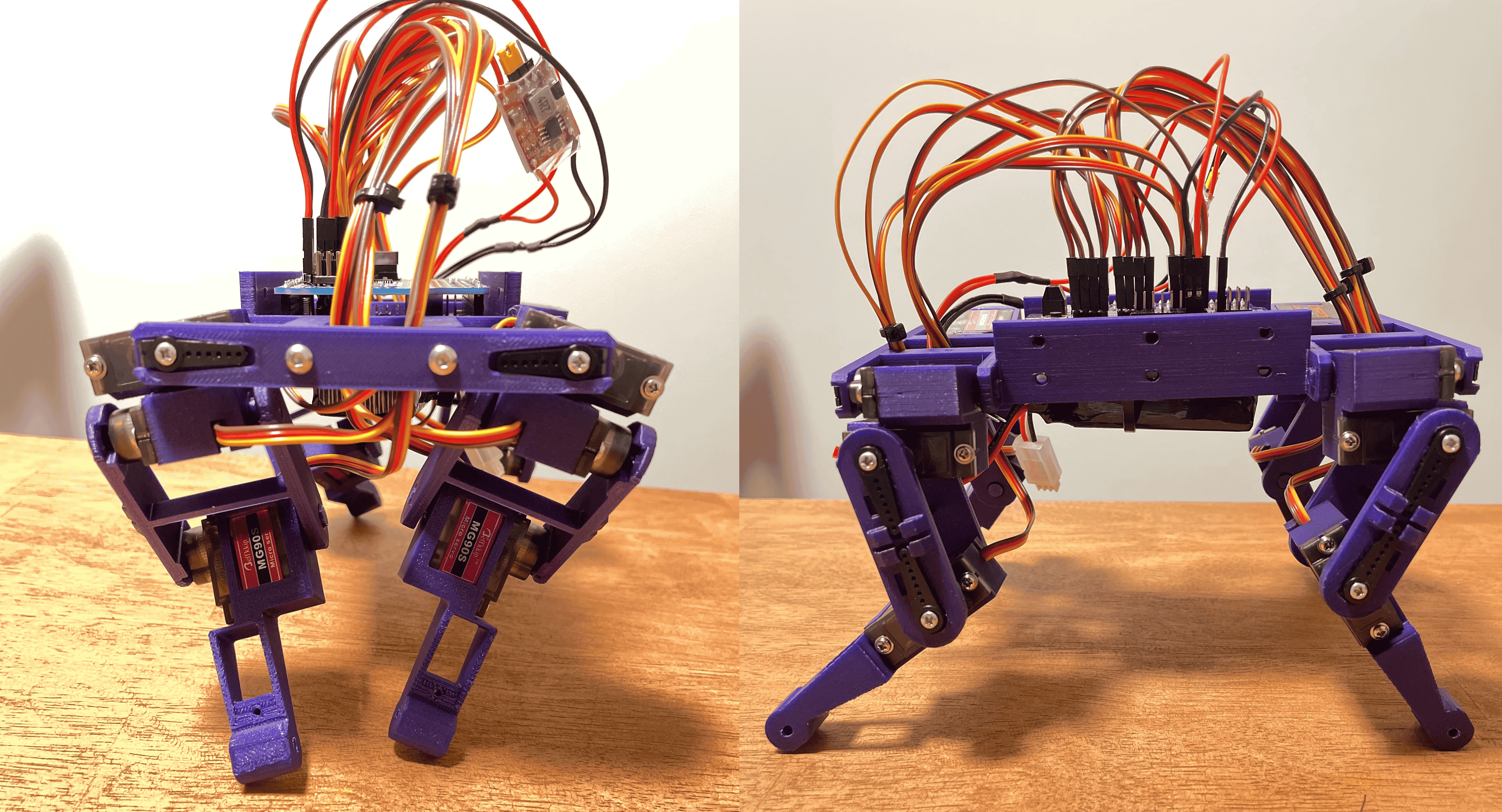}}
\caption{Assembled quadruped robot}
\label{quadruped}
\end{figure}

Each of the four legs of the robot is equipped with three MG90S metal gear micro servos: the Hip Roll Servo, Hip Pitch Servo, and Knee Pitch Servo, providing three degrees of freedom per leg. An Arduino UNO microcontroller serves as the primary control unit of the robot. An Adafruit 16-channel 12-bit Servo Shield is used with the Arduino microcontroller to generate control signals via an I2C communication interface. A 1000 milliampere-hour (mAh) 7.4 Volt Lithium-Polymer battery with a regulated 5 Volt Battery Eliminator Circuit (BEC) powers the robot. The robot is equipped with functions for forward, backward, left, and right movements. Fig. \ref{quadruped} shows the final assembled quadruped robot.

\subsection{Simulation Setup}

The Frozen Lake environment was configured with the default 4$\times$4 grid setting where the agent was trained to navigate, avoid ice holes, and reach the destination. During the training phase, the agent initially explores the environment by taking random actions to learn which grids contain holes. This exploration helps build a Q-table, a probabilistic table that stores the expected rewards for each state-action pair. The agent updates the Q-table using equation \ref{Q-learning equation}, where we set the learning rate, $\alpha$ = 0.1, 
and discount rate, $\gamma$ = 0.99.  

\begin{figure}[htbp]
    \centering
    \begin{subfigure}[t]{0.47\columnwidth}
        \centering
        \includegraphics[width=\linewidth]{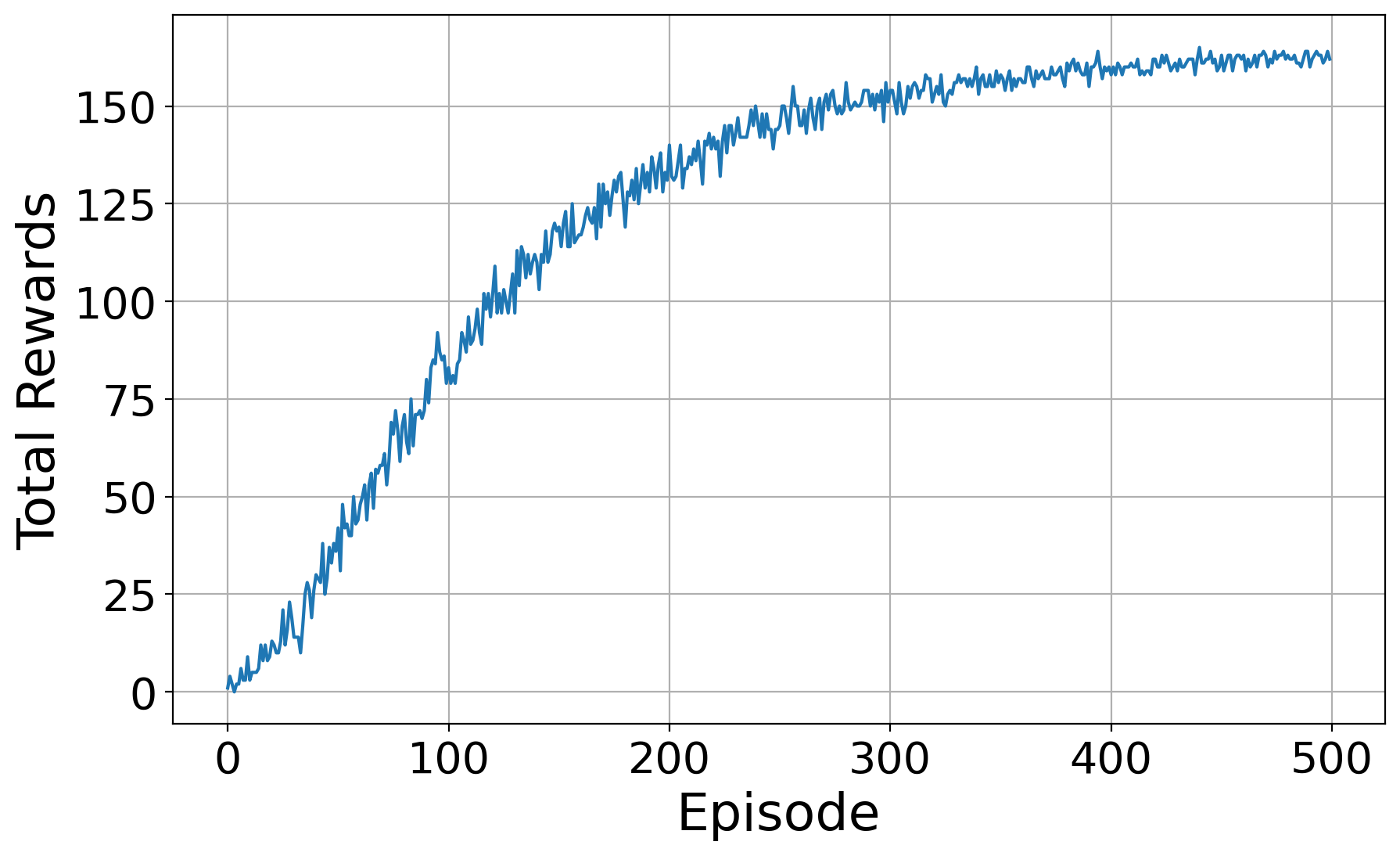}
        \caption{Reward per episode}
        \label{learning_curve}
    \end{subfigure}
    \hspace{0.02\columnwidth} 
    \begin{subfigure}[t]{0.48\columnwidth}
        \centering
        \includegraphics[width=\linewidth]{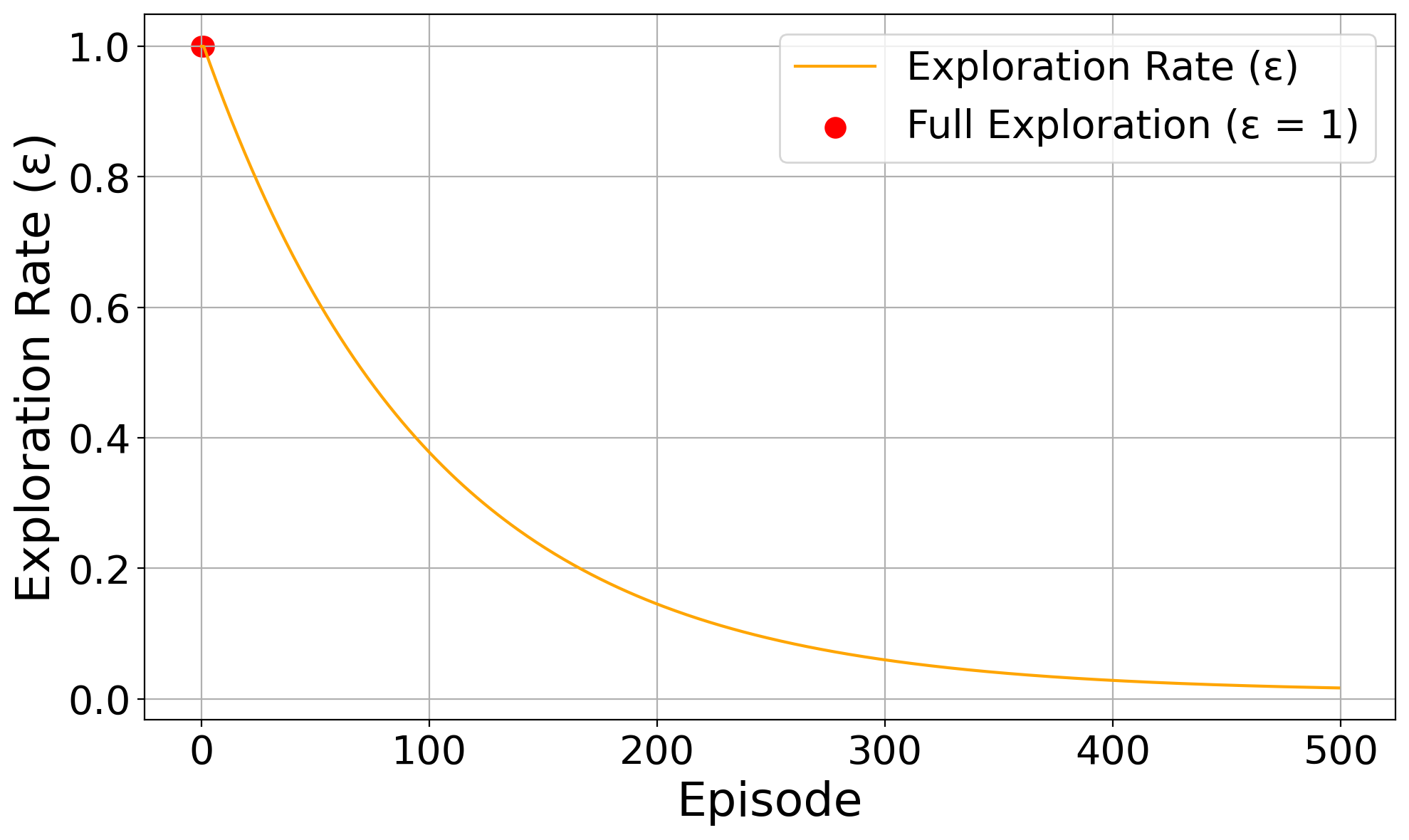}
        \caption{Exploration rate per episode}
        \label{exploration_curve}
    \end{subfigure}
    \caption{Performance of the agent over 500 episodes of training}
    \label{training_plots}
\end{figure}

The training process involved running 500 episodes, each with a maximum of 1000 steps. The number of steps in an episode could vary depending on whether the agent reached the goal or fell into a hole. The agent updates the Q-table after each step and adjusts its policy based on the received rewards. 
As training progresses, the exploration rate $\epsilon$  (which started at 1 $\rightarrow$ fully exploratory), decays over time according the equation: 
\noindent
\begin{equation}
    \epsilon = \epsilon_{min} + (\epsilon_{max} - \epsilon_{min}) \times e^{- decay \hspace{0.1cm} rate \times episode}
\end{equation}

where we set $\epsilon_{min}$ = 0.01, $\epsilon_{max}$ = 1, and decay rate = 0.01. This decay encourages the agent to gradually shift from exploration (random actions) to exploitation (following the learned policy). Fig. \ref{training_plots} shows the learning curve and exploration rate per episode during training.

After training, the optimal policy is extracted from the Q-table, and a sequence of actions is generated. This sequence is used to create a header file ($dataArray.h$ in the code), which contains the actions in a format readable by the Arduino microcontroller of the robot. Based on the header file, the Arduino generates appropriate gaits and enables the robot to navigate based on the trained agent's learned policy. Training the agent and processing the data were conducted in Python using a Jupyter Notebook file, while the implementation of the test results and robot movements were executed in the Arduino codes. This pipeline ensures that the robot can execute the optimal path learned during the Q-learning training phase.

\begin{figure*}[htbp]
\centerline{\includegraphics[width=17.7cm]{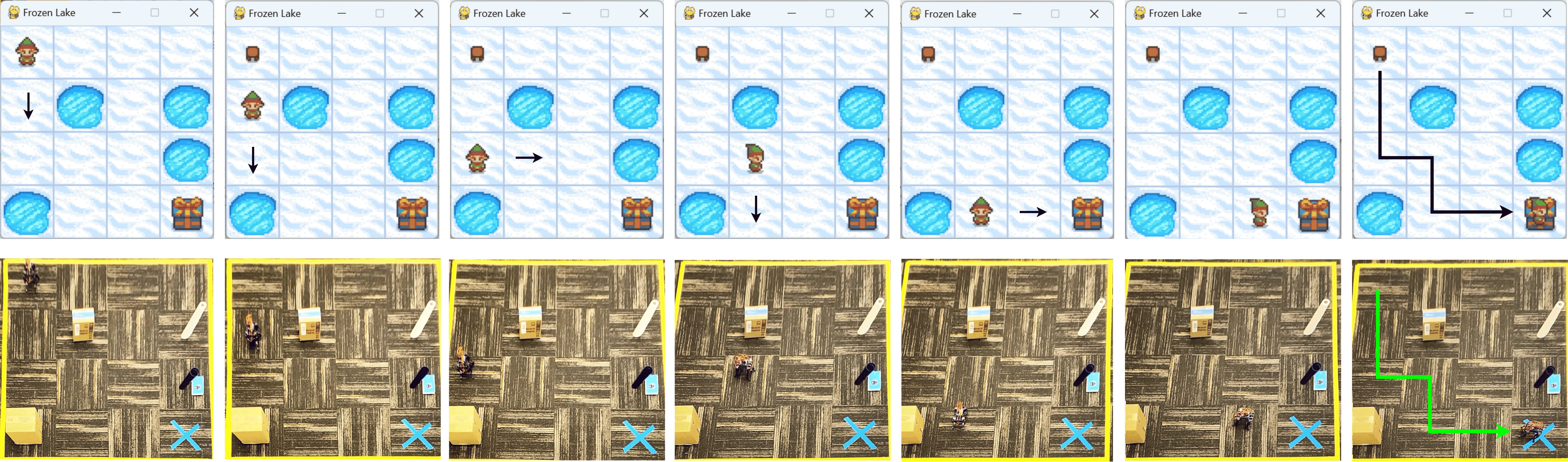}}
\caption{The Sim2Real Transfer Experiment: Demonstration of a quadruped robot navigating in a real grid of size 182 $\times$ 182 cm using a policy learned in Frozen Lake simulation. Four obstacles placed in the grid are customized accordingly in the Frozen Lake simulation. The robot autonomously avoids obstacles without any sensor feedback and reaches the destination using the shortest path.}
\label{sim2real transfer}
\end{figure*}

\begin{figure}[htbp]
\centerline{\includegraphics[width=7cm]{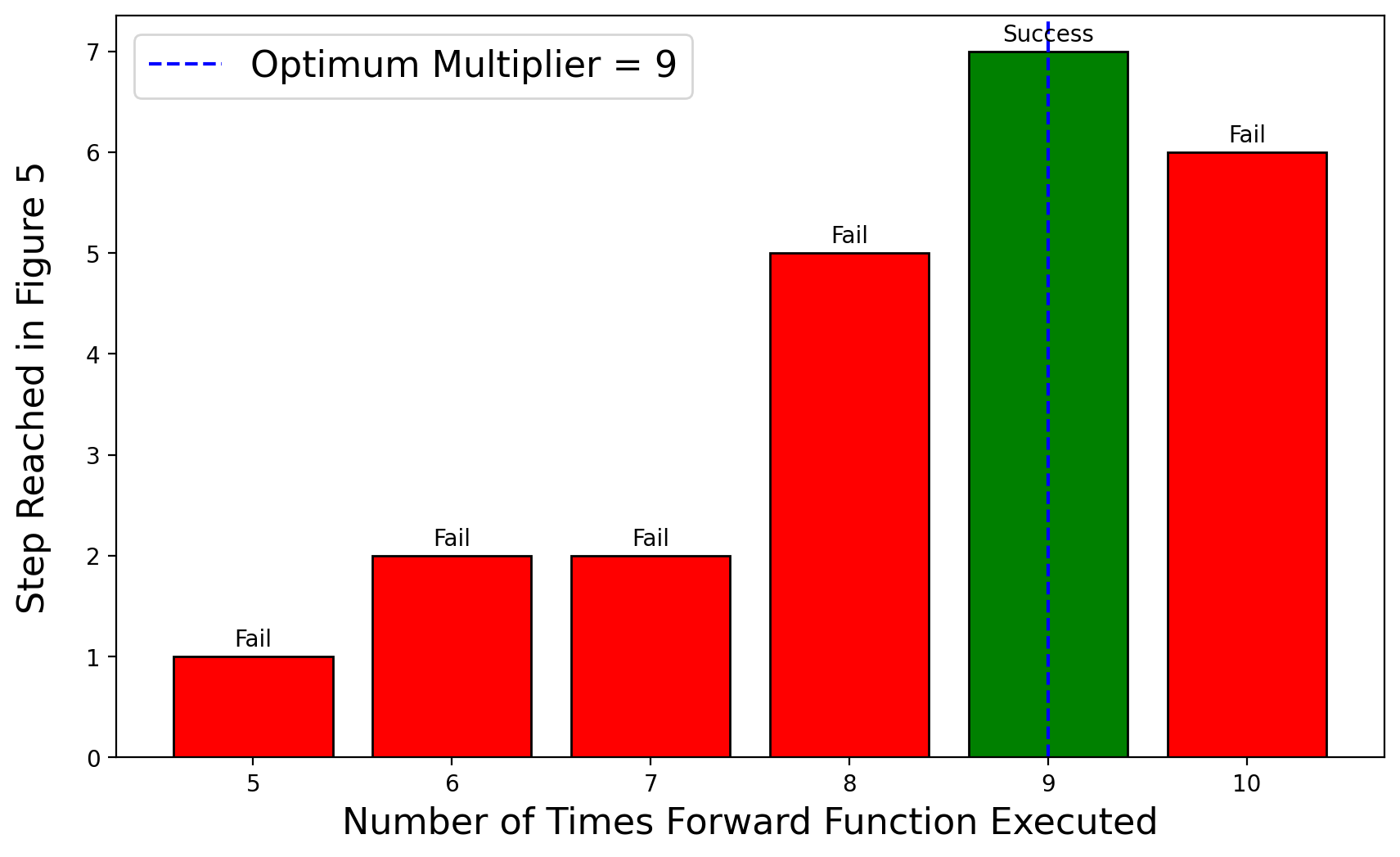}}
\caption{Experiment on optimizing the forward function Multiplier for a 45.5 $\times$ 45.5 cm grid cell.}
\label{multiplier}
\end{figure}

\subsection{Observation and Action}
The Frozen Lake environment consists of 16 discrete observations, each representing a position on the 4$\times$4 grid. The grids are numbered from 0 to 15, and each grid can be represented by coordinates (x, y). The agent can perform one of four actions: 0 (Move left), 1 (Move down), 2 (Move right), and 3 (Move up). For each observation, the Q-table helps determine the best possible action to avoid falling into a hole and to find the shortest route to the goal. For instance, if the agent is at Grid 13, the Q-table may suggest that moving right is the best action. The position of the agent is updated based on the current observation and the chosen action. If the agent's current position is (x, y), the next observation will be updated according to the action taken, as shown in Table \ref{tab}. 
Corresponding to the new observation, the Q-table again updates the action, and a new action is taken.

\begin{table}[htbp]
\caption{Observations for each action in (x, y) coordinates}
\begin{center}
\begin{tabular}{|c|c|c|}
\hline
\textbf{Position} & \textbf{Action} & \textbf{New Observation} \\ \hline
\multirow{4}{*}{$x,\hspace{0.1cm} y$} 
& \textit{Left}  & $x-1, \hspace{0.1cm} y$ \\ \cline{2-3}
& \textit{Right} & $x+1, \hspace{0.1cm} y$ \\ \cline{2-3}
& \textit{Up}    & $x, \hspace{0.1cm} y+1$ \\ \cline{2-3}
& \textit{Down}  & $x, \hspace{0.1cm} y-1$ \\ \hline
\end{tabular}\label{tab}
\end{center}
\end{table}
\noindent

\subsection{Transfer to Reality}

We performed the Sim2Real transfer experiment in a 4$\times$4 grid-like environment at the University of Washington campus, where we placed 4 obstacles exactly in the same positions as the holes in the Frozen Lake environment. The goal of the robot is to avoid obstacles and reach the blue cross mark using the learned policy. Fig. \ref{sim2real transfer} illustrates the 7 steps of the Sim2Real transfer experiment using the Frozen Lake environment.

From Fig. \ref{sim2real transfer}, we can see that the robot successfully transferred the learned policy in Frozen Lake to reality and navigated without relying on any sensor. It avoided the obstacles and also took the shortest path to reach to the destination. The demonstration is also presented in the video.

\subsection{Generalizing Navigation on a Scalable Grid}

To enable the robot to navigate across grids of different sizes, we conducted an experiment to introduce a multiplying factor for the forward function of the robot. This factor intends to ensure consistent movement from the center of one grid cell to the center of the adjacent cell. In the experimental setup, each grid cell measures approximately 45.5 $\times$ 45.5 cm. The investigation involved testing with different forward function multipliers ranging from 5 to 10, with each multiplier executed five times to determine the optimal value. The outcomes of these trials are summarized in Fig. \ref{multiplier}. The results demonstrate that a multiplying factor of 9 consistently led to successful navigation, allowing the robot to reach the destination in all test cases. Adjusting the multiplying factor will enable the robot to adapt to grids of various scales, thereby generalizing navigation in different grid environments.

\section{Conclusion and Future Work}\label{conclusion}

In this paper, we presented an accessible approach for Sim2Real transfer in sensor-independent robot navigation using the Gymnasium Frozen Lake environment and a 12-DOF quadruped robot. This work contributes to the robotics field by demonstrating that a simplified grid-based simulation environment, coupled with a Q-learning algorithm, can effectively be translated to real-world applications without relying on sensors or additional real-world training. We demonstrate that the robot is capable of navigating a 4$\times$4 grid, avoiding obstacles, and reaching its target based on the learned policy without any feedback. We argue that this is one of the most straightforward Sim2Real approaches for grid-based environments and significant for educational purposes, hobbyist projects, and early-career researchers who require affordable and open-source solutions to explore RL applications in robotics.

At its current stage, our Sim2Real approach has some limitations, such as a lack of comparisons with other similar simulation environments and limited tests confined to (i) specific Frozen Lake configurations, and (ii) single robot setup. To address these, our future work will include running additional experiments with a variety of Frozen Lake configurations and implementing them in similar real-world settings. We also plan to perform the Sim2Real transfer experiment on a variety of robot platforms, including wheeled robots, to evaluate the adaptability of the approach. In addition, we will explore the impact of the generalization factor on policy transferability across different grid environments.

\end{document}